%% file: emnlp-ijcnlp-2019.tex
\documentclass[11pt,a4paper]{article}
\usepackage[hyperref]{emnlp-ijcnlp-2019}
\usepackage{times}
\usepackage{latexsym}
\usepackage{url}
\usepackage{graphicx}
\aclfinalcopy 

\title{An Empirical Study of Factors Affecting Language-Independent Models}

\author{Xiaotong Liu \\
  IBM Research-Almaden \\
  {\tt \normalsize{xiaotong.liu@ibm.com}} \\\And
  Yingbei Tong \\
  IBM Research-Almaden \\
  {\tt \normalsize{yingbei.tong@ibm.com}} \\ \\
  \textbf{Rama Akkiraju} \\
  IBM Research-Almaden \\
  {\tt \normalsize{akkiraju@us.ibm.com}} \\ \And
  Anbang Xu \\
  IBM Research-Almaden \\
  {\tt \normalsize{anbangxu@us.ibm.com}}
}

\date{}

\begin{document}
\maketitle
\begin{abstract}
Scaling existing applications and solutions to multiple human languages has traditionally proven to be difficult, mainly due to the language-dependent nature of preprocessing and feature engineering techniques employed in traditional approaches. In this work, we empirically investigate the factors affecting language-independent models built with multilingual representations, including task type, language set and data resource. On two most representative NLP tasks --- sentence classification and sequence labeling, we show that language-independent models can be comparable to or even outperforms the models trained using monolingual data, and they are generally more effective on sentence classification. We experiment language-independent models with many different languages and show that they are more suitable for typologically similar languages. We also explore the effects of different data sizes when training and testing language-independent models, and demonstrate that they are not only suitable for high-resource languages, but also very effective in low-resource languages.
\end{abstract}

\input{intro}
\input{relatedwork.tex}
\input{method}

\input{evaluation}
\input{conclusion}

\bibliography{emnlp-ijcnlp-2019}
\bibliographystyle{acl_natbib}

\end{document}

%% file: intro.tex
\section{Introduction}

In today's globalized world, companies need to be able to understand and analyze what is being said out there, about them, their products, services, or their competitors, regardless of the human language used.  Many organizations have spent tremendous resources to develop cognitive applications and services for dealing with customers in different countries. For example, cognitive systems may use machine learning techniques to process input messages or statements to determine their meaning and to provide associated confidence scores based on knowledge acquired by the cognitive system. Typically, the use of such cognitive systems requires training individual natural language understanding models in a specific human language. For example, a tone analyzer model can be built to predict tones from English conversations~\cite{liu2018voice}, but such model would not work effectively with other languages.  While translation techniques can be applied to translate data from an existing language to another language, human translation is labor-intensive and time-consuming, and machine translation can be costly and unreliable. As a result, attempts to scale existing applications to multiple human languages has traditionally proven to be difficult, mainly due to the language-dependent nature of preprocessing and feature engineering techniques employed in traditional approaches~\cite{akkiraju2018characterizing}.

In this work, we empirically investigate the feasibility of multilingual representations to build \textit{language-independent models}, which can be trained with data from multiple \textit{source languages} and then serve multiple \textit{target languages} (target languages can be different from source languages). We explore this question using a unified language model \textit{Multilingual BERT}~\cite{devlin2019bert}, which is pre-trained on the combination of monolingual Wikipedia corpora from 104 languages. Through a series of experiments on multiple task types, language sets and data resources, we contribute empirical findings of how factors affect language-independent models:
\begin{itemize}
	\item \textbf{Task Type.} We analyze and compare language-independent models on two most representative NLP tasks: sentence classification and sequence labeling. On both tasks, we show that language-independent models can be comparable to or even outperform the models trained using monolingual data. Language-independent models are generally more effective on sentence classification.
	\item \textbf{Language Set.} Theoretically language-independent models can be trained using any language set, and be used to make predictions in any language. Through training and testing language-independent models with many different languages, we show that they are more suitable for typologically similar languages. 
	\item \textbf{Data Resource.} We explore the effects of different data sizes when training language-independent models. We demonstrate that language-independent models are not only suitable for high-resource languages, but also very effective in low-resource languages.
\end{itemize}

We derive insights from our experiments to facilitate the development and customization of natural language understanding models and solutions in new languages. First of all, it can be used to solve the \textit{cold-start} problem, where no initial model is available for a new target language, when building such models from scratch is costly. Secondly, it largely saves the cost and time for acquiring annotated data of a new target language by reusing data already annotated in previously supported languages. Thirdly, it simplifies the deployment process of a new model and save the efforts for simultaneously maintaining multiple monolingual models in a production setting. Our annotated data for low-resource languages will be made publicly available.

%% file: relatedwork.tex
\section{Related Works}

Multilingual representation learning has been an active area of research, starting from word embeddings alignment that uses small dictionaries to align word representations from different languages~\cite{mikolov2013exploiting}. Research by~\cite{faruqui2014improving} has demonstrated that multilingual representations can be leveraged to improve the quality of monolingual representations. An unsupervised learning method has been proposed by~\cite{conneau2017word} to align multilingual word embeddings without parallel data. In addition to word embedding alignment, aligning sentence representations from multiple languages has also been studied in machine translation, on both supervised learning~\cite{johnson2017google, artetxe2018massively} and unsupervised learning~\cite{lample2017unsupervised, artetxe2017unsupervised}. However, most of these approaches focus on pairwise multilingual representation learning. In this work, we empirically investigate the impact of multilingual representations learned from a large number of languages on tasks that involves more languages than a certain language pair.

Our work builds on top of recent advances in pre-trained language modeling. ELMo~\cite{peters2018deep} extracts context-sensitive features from a bidirectional LSTM language model and provides additional features for a task-specific architecture. ULMFiT~\cite{howard2018universal} advocates discriminative fine-tuning and slanted triangular learning rates to stabilize the fine-tuning process with respect to end tasks. OpenAI GPT~\cite{radford2018improving} builds on multi-layer transformer~\cite{vaswani2017attention} decoders instead of LSTM to achieve effective transfer while requiring minimal changes to the model architecture. Recently, BERT~\cite{devlin2019bert} uses bidirectional transformer encoders to pre-train a large corpus, and fine-tunes the pre-trained model that requires almost no specific architecture for each end task. 
In this work, we leverage the multilingual representations learned from multilingual BERT~\cite{devlin2019bert} to build models that can scale to many languages. 

%% file: method.tex
\section{Language-Independent Model}

In this section, we describe the motivation of language-independent models, and how to create such models via multilingual representation learning and fine-tuning.

\subsection{One Model, Many Languages}

\begin{figure}
  \includegraphics[width=\linewidth]{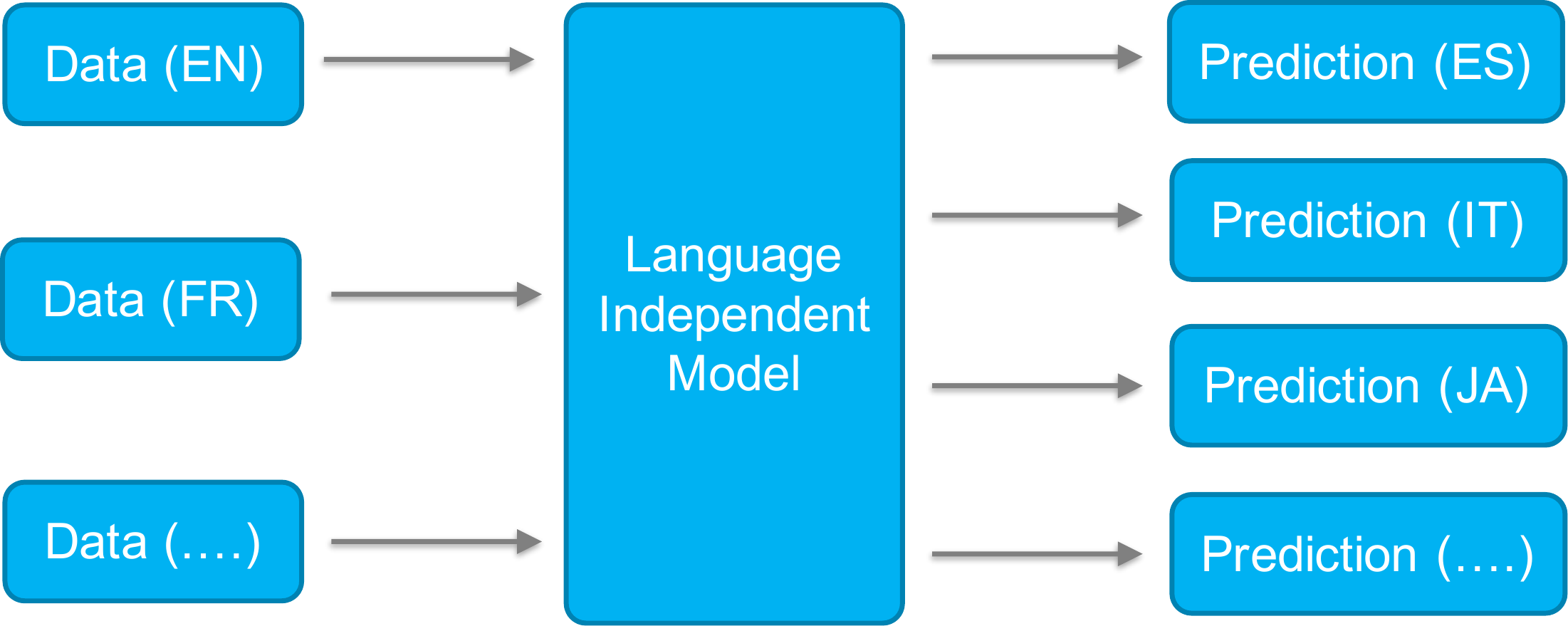}
  \caption{A conceptual example of a Language-Independent Model (LIM). The target languages to serve in production do not necessarily depend on which source languages were used in training. For instance, an LIM can be trained using annotated data from the source languages such as English (EN) and French (FR), and then serve in the target languages including Spanish (ES), Italian (IT), Japanese(JA) and so on. } \label{figure:lim}
\end{figure}

To scale our efforts to support the diversity of people in the world, it is important to build and customize machine learning models for many different languages in various NLP tasks. For each target language, however, this often requires going through the whole lifecycle of data collection, data cleansing, data annotation, data storage, feature creation and selection, machine learning model training, model validation, benchmarking and deployment of these models as services in production~\cite{akkiraju2018characterizing}. It easily becomes overwhelming as the number of target languages increases. To address this problem, we advocate to build one model for all target languages together, which we called a \textit{Language-Independent Model (LIM)}, as the target languages to serve in production do not necessarily depend on which source languages were used in training. Figure~\ref{figure:lim} shows a conceptual example: an LIM can be trained using annotated data from the source languages such as English (EN) and French (FR), and then serve in the target languages including Spanish (ES), Italian (IT), Japanese(JA), which are different from the source languages. This not only accelerates the enablement of a new language by reusing data already annotated in previously supported languages, but also simplifies the deployment process and save efforts for maintaining multiple monolingual models in production.

\subsection{Multilingual Representation Learning with BERT}

The basis for building LIMs lies in learning a representation that can feature multiple languages. Among the recent significant advances in deep contextualized representation learning for natural language understanding, BERT~\cite{devlin2019bert} stands out as its pre-training process naturally supports multilingual representation learning. Specifically, multilingual BERT was pre-trained on the Wikipedia pages (excluding user and talk pages) of 104 languages with a 110K shared WordPiece~\cite{wu2016google} vocabulary. It is a 12-layer, 768-hidden, 12-head transformer model~\cite{vaswani2017attention} with 110M parameters. To alleviate the bias towards high-resource languages such as English, data from high-resource languages were under-sampled and those from low-resource languages were over-sampled. The pre-training of multilingual BERT does not use any marker denoting the input language, and does not rely on parallel corpus to explicitly encourage translation-equivalent pairs to have similar representations. 

\subsection{Fine-Tuning Multilingual BERT for End Tasks}

\begin{figure}
  \includegraphics[width=\linewidth]{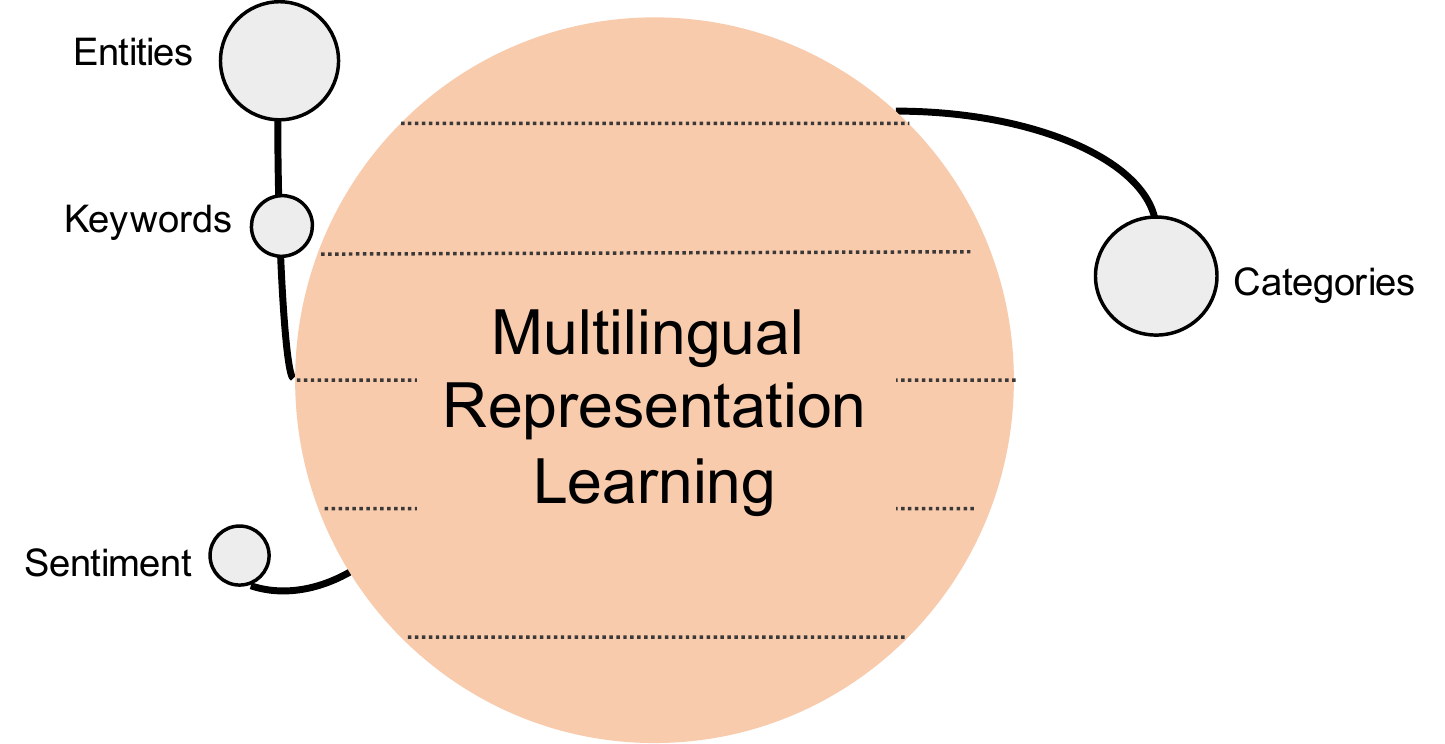}
  \caption{An illustration of generalized multilingual representation learning for different NLP tasks.} \label{figure:bert}
\end{figure}

The multilingual representations learned with BERT can be generalized for many natural language understanding tasks such as Sentiment Analysis, Named Entity Recognition, Categorization, and so on (as illustrated in Figure~\ref{figure:bert}). The input representation of multilingual BERT is a sequence of tokens in any language, which may be a single sentence or two sentences packed together. The input representation of each token is constructed as the sum of the corresponding token, segment, and position embeddings. For sentence classification tasks, the first token of each sequence is a special classification embedding ([CLS]) and its final hidden state will be used as the aggregate representation of the whole sequence. For sequence labeling tasks, the final hidden state of each token will encode its contextualized representation with respect to the whole sequence. To fine-tune multilingual BERT, a classification layer is added on top of the final representation layer, and the probabilities of all label classes are computed with a standard softmax. The parameters of multilingual BERT and the classification layer are fine-tuned jointly to maximize the log-probability of the correct label. The labeled data of end tasks are shuffled across different languages when fine-tuning multilingual BERT.

%% file: evaluation.tex
\section{Experiments}

The effects of LIMs can be affected by at least three factors: task type, language set and data resource. In this section, we empirically investigate the effects of these factors on the performance of LIMs.

\subsection{Factor Characterization}

\paragraph{Task Type} We explore whether LIMs are equally effective across different end tasks. For the scope of this paper, we consider sentence classification and sequence labeling as two of the most popular NLP tasks. In particular, we select and compare two representative tasks: Sentiment Analysis and Named Entity Recognition (NER). Sentiment Analysis represents a typical sentence classification task, while NER is a popular sequence labeling task. 

\paragraph{Language Set} While theoretically an LIM can be trained using any language set, and be used to make predictions in any language, multilingual representations may not be equally effective across different languages~\cite{gerz2018relation}. For instance, it has been shown that a multilingual word embedding alignment between English and Chinese is much more difficult to learn than that between English and Spanish~\cite{conneau2017word}. We explore many different languages when training and testing LIMs.

\paragraph{Data Resource} For high-resource languages, the annotated data can be of different sizes; for low-resource languages, large amounts of data do not often exist~\cite{kasai2019low}. We explore the effects of different data sizes when training and testing LIMs.

\subsection{Case Study on Sentiment Analysis} \label{section:model}

We take Sentiment Analysis as a 3-class classification problem: given a sentence $s$ in a target language $T$, which consists of a series of words: $\{w_1, ... , w_m\}$, predict the sentiment polarity $y \in \{positive, neutral, negative\}$. 

For this case study, we consider 7 high-resource languages: English, Spanish, Italian, Brazilian Portuguese, Dutch, Japanese and Chinese, covering both western and eastern languages. The high-resource training set consists of 770K data points --- 230K English, and 90K each in other 6 languages; the test set contain both public available test data and high quality in-house test data --- 630K English, 10K Spanish, 57K Japanese, 10K Chinese and 15K French. Meanwhile, we collect 5K data points each in 5 languages: Danish, Swedish, Norwegian, Russian, and Turkish, which are considered as low-resource languages in our experiments. We use 4K as training set and 1K as test set for each low-resource language. 

We randomly split 1/10 from the training set as the development set for model selection and the rest for model training (i.e., fine-tuning the parameters of Multilingual BERT and the sentence classification layer). Following original BERT fine-tuning~\cite{devlin2019bert}, we fine-tune the multilingual BERT with the following parameter choices: (1) batch size: 16, 32; (2) learning rate: 5e-5, 3e-5, 2e-5; (3) number of epochs: 3, 4. The model of 32 batch size, 2e-5 learning rate and 4 epochs was selected as the best model based on its performance on the development set. We denote the LIM for Sentiment Analysis trained with high-resource languages as \textbf{LIM-H}, and the LIM trained with the mix of high-resource and low-resource languages as \textbf{LIM-M}.

\subsubsection{Results on High-Resource Languages}

For high-resource languages, we compare \textbf{LIM-H} with the following methods:

\begin{itemize}
	\item \textbf{CNN}~\cite{kim2014convolutional} is a convolutional neural networks (CNN) trained on top of pre-trained word vectors for sentence-level classification tasks. We use this method to train monolingual Sentiment Analysis models as a baseline because of its popularity and simple implementation for reproducibility. 
	\item \textbf{ULMFiT}~\cite{howard2018universal} is a recent generative pretrained language model with task-specific fine-tuning. We follow ULMFiT by adopting discriminative fine-tuning and slanted triangular learning rates to stabilize the fine-tuning process and create monolingual Sentiment Analysis models.
	\item \textbf{Monolingual-BERT}. We trained monolingual Sentiment Analysis models by fine-tuning BERT with monolingual datasets for every language, respectively. For example, a \textit{Chinese-only BERT model} refers to the BERT model fine-tuned using Chinese-only annotated data for Sentiment Analysis.
\end{itemize}

In Table~\ref{table:sentiment_1}, we report the accuracy results of Sentiment Analysis on English and Spanish across various models. We get a significant boost in performance of 7.4\% than CNN, and 3.2\% than ULMFiT in English. As for Spanish, we outperform the previous methods by 4.5\% and 2.3\% respectively. 

\begin{table} [!htb]
\center
\begin{tabular}{l|ccc}
Language & CNN & ULMFiT & LIM-H  \\
\hline
English & 72.1 & 76.3& \textbf{79.5} \\
Spanish & 69.4 & 71.6 & \textbf{73.9} \\
\end{tabular}
\caption{Accuracy results of Sentiment Analysis on English and Spanish across various models.}   \label{table:sentiment_1}
\end{table}

Furthermore, we show that our method is able to compete with the monolingual BERT models on Sentiment Analysis in Table~\ref{table:sentiment_2}. By leveraging data from non-native languages, our LIM outperforms the English-only BERT model by 1.8\% and the Japanese-only BERT model by 0.7\%, but falls behind the Chinese-only BERT model by 1.2\%. It should be noted that BERT specifically pre-trained the Chinese-only model to account for its unique character tokenization. Therefore, it is still very encouraging to see that our LIM is comparable to a specially customized monolingual BERT model. 

\begin{table} [!htb]
\center
\begin{tabular}{l|cc}
Language & Monolingual-BERT & LIM-H \\
\hline
English & 77.7 & \textbf{79.5}\\
Japanese & 78.0 & \textbf{78.7}\\
Chinese & \textbf{74.5} & 73.3
\end{tabular}
\caption{Accuracy results of Sentiment Analysis on English, Japanese and Chinese between monolingual BERT and LIM-H.}   \label{table:sentiment_2}
\end{table}

In Table~\ref{table:sentiment_3}, we evaluate the impact of LIM on Sentiment Analysis via zero-shot transfer learning. When we do not include any French annotated data for training, we can still obtain a significant improvement of 5.7\% over the monolingual CNN model trained using French annotated data. 

\begin{table} [!htb]
\center
\begin{tabular}{l|cc}
Language & CNN  & LIM-H \\
\hline
French & 54.0 & \textbf{59.7} 
\end{tabular}
\caption{Accuracy results of Sentiment Analysis on French between CNN and LIM-H. This demonstrates a \textit{zero-shot transfer learning} case for LIM-H as it does not involve any French annotated data when training the model.}   \label{table:sentiment_3}
\end{table}

\subsubsection{Results on Low-Resource Languages}

For low-resource languages, we compare both \textbf{LIM-H} and \textbf{LIM-M} in Table~\ref{table:sentiment_4}. LIM-H demonstrates the effects of zero-shot transfer learning on low-resource languages, with an average of 60\% accuracy. Since we do not use any low-resource training data in LIM-H, this shows that LIM can be used to address the \textit{cold-start} problem, where no initial model is available for a new target low-resource language, when building such models from scratch is costly. Furthermore, LIM-M demonstrates how much improvement a LIM can gain by adding only a small amount of data in low-resource languages. In particular, by adding 4K annotated data in each low-resource language, we obtain an average of 11\% improvement. This largely saves the cost and time for acquiring annotated data of a new target low-resource language by transferring the knowledge learned from a larger amount of annotated data available in high-resource languages.

\begin{table} [!htb]
\center
\begin{tabular}{l|cc}
Language & LIM-H  & LIM-M\\
\hline
Danish & 62.5 & \textbf{69.2} \\
Swedish & 56.8 & \textbf{68.6} \\
Norwegian & 62.0 & \textbf{70.3} \\
Russian & 62.1 & \textbf{75.8} \\
Turkish & 56.8 & \textbf{69.1} 
\end{tabular}
\caption{Accuracy results of Sentiment Analysis on low-resource languages. We compare the performance of zero-shot transfer learning in LIM=H (without any annotated data from the target languages) and low-resource transfer training in LIM-M (only 4K annotated data from the target languages were used in training).}   \label{table:sentiment_4}
\end{table}

\subsection{Case Study on Named Entity Recognition}

Given a sentence $s$ in a target language $T$, which consists of a series of words: $\{w_1, ... , w_m\}$, NER outputs a sequence of labels $\{l_1, ... , l_m\}$, with respect to the named entity type $e \in $~\{\textit{Person, Location, Organization, Date, Time, JobTitle, Duration, Facility, GeographicFeature, Measure, Ordinal, Money}\}. This is much more fine-grained and complex than the traditional CoNLL NER task that only considers 4 entity types~\cite{TjongKimSang:2002, TjongKimSang:2003}. We follow the \textit{Inside–-outside–-beginning (IOB2)} tagging format~\cite{ramshaw1999text}: a $B$-prefix means that the tag is the beginning of a chunk, an $I$-prefix indicates that the tag is inside a chunk, and an $O$ tag represents that a token belongs to no chunk.

We build an LIM for NER with annotated data in 3 languages: French, Italian and German. The training set consists of 679K data points (148K in French, 470K in Italian and 61K in German). We randomly split 1/10 from the training set as the development set for model selection and the rest for model training (i.e., fine-tuning the parameters of Multilingual BERT and the sequence labeling layer). We selected the best model of 32 batch size, 2e-5 learning rate and 3 epochs, after fine-tuning with different parameters (described in Section~\ref{section:model}) on the development set.

\subsubsection{Compared Methods}

We compare \textbf{LIM} with the following methods:

\begin{itemize}
	\item \textbf{BiLSTM+CRF}~\cite{lample2016neural} is a bidirectional LSTM with a sequential conditional random field above it. We use this method to train monolingual NER models as a baseline because it has been effective and widely used on sequence labeling tasks.  
	\item \textbf{FLAIR}~\cite{akbik2019flair} is one of the latest NLP frameworks that achieved state-of-the-art for sequence labeling tasks. It models words as sequence of characters and leverages contextual string embeddings produced from a trained character language model~\cite{akbik2018contextual}. We adopt the pre-trained multilingual FLAIR embedding to build multilingual NER models using the FLAIR framework.
\end{itemize}

\subsubsection{Results}

We evaluate the models on high quality in-house benchmark datasets for NER in various languages including French (3870 entities), Italian (3776 entities), and German (5023 entities)\footnote{We refer to the number of entities instead of data points as one data point can contain multiple entities.}. 

First of all, we report the F-measure results of NER on French, Italian and German. Regarding French, we reach a significant improvement in performance of 9.9\% than BiLSTM+CRF, and 7.1\% than FLAIR. Similarly, on German, we outperform the previous methods by 6.1\% and 2.4\% respectively. Our LIM approach is comparable to BiLSTM+CRF and outperforms FLAIR by 3.5\% on Italian.

\begin{table} [!htb]
\center
\begin{tabular}{l|ccc}
Language & BiLSTM+CRF & FLAIR & LIM \\
\hline
French & 68.0 & 70.8 & \textbf{77.9}  \\
Italian & 71.5 & 68.0 & 71.5  \\
German  & 64.5 & 68.2 & \textbf{70.6}  \\
\end{tabular}
\caption{F-measure results of NER on French, Italian and German. The BiLSTM-CRF models were trained using monolingual data in each language respectively. The FLAIR and LIM models were trained using the concatenation of French, Italian and German annotated data.}   \label{table:ner_1}
\end{table}

Secondly, we evaluate the effects of our LIM approach for zero-shot transfer learning on NER. We trained another FLAIR and LIM using only the concatenation of French and Italian annotated data while excluding German annotated data. Table~\ref{table:ner_2} shows that our LIM method is able to retain the performance of 58.6\% while FLAIR drops to 20.3\%. demonstrates shows the power of our LIM method in accelerating the development of models for a new language where no annotated data is available.

\begin{table} [!htb]
\center
\begin{tabular}{l|cc}
Language & FLAIR & LIM \\
\hline
German & 20.3 & \textbf{58.6}  \\
\end{tabular}
\caption{F-measure results of NER on German (zero-shot transfer learning). The FLAIR and LIM models were trained using the concatenation of French and Italian annotated data, while German annotated data was excluded.}   \label{table:ner_2}
\end{table}

\subsection{Discussion}

\paragraph{Task Type} While the results demonstrate the effectiveness of LIMs on two most representative NLP tasks, we found that LIMs are generally more effective on a sentence classification task than a sequence labeling task, particularly for zero-shot transfer learning. For example, LIM outperforms the corresponding baseline on Sentiment Analysis (Table~\ref{table:sentiment_3}), but falls behind the corresponding baseline on NER (Table~\ref{table:ner_1} and~\ref{table:ner_2}), when no annotated data from the target language was used in model training.

\paragraph{Language Set} Powered by the multilingual representations learned in pre-trained BERT, LIMs seem more suitable for typologically similar languages. For instance, the LIM-H is not as good as the model trained using Chinese-only BERT on Sentiment Analysis, though the difference is relatively small (Table~\ref{table:sentiment_2}). This is consistent with the findings from multilingual representation learning using word embeddings~\cite{conneau2017word}.

\paragraph{Data Resource} Language-independent models are not only suitable for high-resource languages, but also very effective in low-resource languages. In particular, adding a relatively small amount of low-resource training data can result in a significant improvement of performance (Table~\ref{table:sentiment_4}).

\paragraph{Implications} These insights bring unique values to the development and customization of natural language understanding models and solutions in new languages. First of all, it can be used to solve the cold-start problem, where no initial model is available for a new target language, when building such models from scratch is costly. Secondly, it largely saves the cost and time for acquiring annotated data of a new target language by reusing data already annotated in previously supported languages. Thirdly, it simplifies the deployment process of a new model and save the efforts for simultaneously maintaining multiple monolingual models in a production setting.

%% file: conclusion.tex
\section{Conclusion and Future Work}

As the use of machine learning becomes more pervasive all over the world, people speaking different languages will come to expect seamless and customized experience of their own. Building a language independent model can accelerate the enablement of machine learning and cognitive solutions in new languages at a large scale. We demonstrate the power of this language-independent modeling approach through a series of experiments on multiple task types, language sets and data resources. Our annotated data for low-resource languages will be made publicly available. We hope that the insights gained from these experiments will help researchers and practitioners develop solutions and tools that enable better scalability, integration and operations in many other languages. In future, we will continue to explore the effects of different combinations of languages with respect to various end tasks. Besides, we plan to extend the studies to more NLP tasks, and investigate the feasibility of multi-task learning for building a task and language independent framework.

%% file: emnlp-ijcnlp-2019.bbl
\begin{thebibliography}{24}
\expandafter\ifx\csname natexlab\endcsname\relax\def\natexlab#1{#1}\fi

\bibitem[{Akbik et~al.(2019)Akbik, Bergmann, Blythe, Rasul, Schweter, and
  Vollgraf}]{akbik2019flair}
Alan Akbik, Tanja Bergmann, Duncan Blythe, Kashif Rasul, Stefan Schweter, and
  Roland Vollgraf. 2019.
\newblock Flair: An easy-to-use framework for state-of-the-art nlp.
\newblock In \emph{Proceedings of the 2019 Conference of the North American
  Chapter of the Association for Computational Linguistics (Demonstrations)},
  pages 54--59.

\bibitem[{Akbik et~al.(2018)Akbik, Blythe, and Vollgraf}]{akbik2018contextual}
Alan Akbik, Duncan Blythe, and Roland Vollgraf. 2018.
\newblock Contextual string embeddings for sequence labeling.
\newblock In \emph{Proceedings of the 27th International Conference on
  Computational Linguistics}, pages 1638--1649.

\bibitem[{Akkiraju et~al.(2018)Akkiraju, Sinha, Xu, Mahmud, Gundecha, Liu, Liu,
  and Schumacher}]{akkiraju2018characterizing}
Rama Akkiraju, Vibha Sinha, Anbang Xu, Jalal Mahmud, Pritam Gundecha, Zhe Liu,
  Xiaotong Liu, and John Schumacher. 2018.
\newblock Characterizing machine learning process: A maturity framework.
\newblock \emph{arXiv preprint arXiv:1811.04871}.

\bibitem[{Artetxe et~al.(2017)Artetxe, Labaka, Agirre, and
  Cho}]{artetxe2017unsupervised}
Mikel Artetxe, Gorka Labaka, Eneko Agirre, and Kyunghyun Cho. 2017.
\newblock Unsupervised neural machine translation.
\newblock \emph{arXiv preprint arXiv:1710.11041}.

\bibitem[{Artetxe and Schwenk(2018)}]{artetxe2018massively}
Mikel Artetxe and Holger Schwenk. 2018.
\newblock Massively multilingual sentence embeddings for zero-shot
  cross-lingual transfer and beyond.
\newblock \emph{arXiv preprint arXiv:1812.10464}.

\bibitem[{Conneau et~al.(2017)Conneau, Lample, Ranzato, Denoyer, and
  J{\'e}gou}]{conneau2017word}
Alexis Conneau, Guillaume Lample, Marc'Aurelio Ranzato, Ludovic Denoyer, and
  Herv{\'e} J{\'e}gou. 2017.
\newblock Word translation without parallel data.
\newblock \emph{arXiv preprint arXiv:1710.04087}.

\bibitem[{Devlin et~al.(2019)Devlin, Chang, Lee, and
  Toutanova}]{devlin2019bert}
Jacob Devlin, Ming-Wei Chang, Kenton Lee, and Kristina Toutanova. 2019.
\newblock Bert: Pre-training of deep bidirectional transformers for language
  understanding.
\newblock In \emph{Proceedings of the 2019 Conference of the North American
  Chapter of the Association for Computational Linguistics: Human Language
  Technologies, Volume 1 (Long and Short Papers)}, pages 4171--4186.

\bibitem[{Faruqui and Dyer(2014)}]{faruqui2014improving}
Manaal Faruqui and Chris Dyer. 2014.
\newblock Improving vector space word representations using multilingual
  correlation.
\newblock In \emph{Proceedings of the 14th Conference of the European Chapter
  of the Association for Computational Linguistics}, pages 462--471.

\bibitem[{Gerz et~al.(2018)Gerz, Vuli{\'c}, Ponti, Reichart, and
  Korhonen}]{gerz2018relation}
Daniela Gerz, Ivan Vuli{\'c}, Edoardo~Maria Ponti, Roi Reichart, and Anna
  Korhonen. 2018.
\newblock On the relation between linguistic typology and (limitations of)
  multilingual language modeling.
\newblock In \emph{Proceedings of the 2018 Conference on Empirical Methods in
  Natural Language Processing}, pages 316--327.

\bibitem[{Howard and Ruder(2018)}]{howard2018universal}
Jeremy Howard and Sebastian Ruder. 2018.
\newblock Universal language model fine-tuning for text classification.
\newblock In \emph{Proceedings of the 56th Annual Meeting of the Association
  for Computational Linguistics (Volume 1: Long Papers)}, pages 328--339.

\bibitem[{Johnson et~al.(2017)Johnson, Schuster, Le, Krikun, Wu, Chen, Thorat,
  Vi{\'e}gas, Wattenberg, Corrado et~al.}]{johnson2017google}
Melvin Johnson, Mike Schuster, Quoc~V Le, Maxim Krikun, Yonghui Wu, Zhifeng
  Chen, Nikhil Thorat, Fernanda Vi{\'e}gas, Martin Wattenberg, Greg Corrado,
  et~al. 2017.
\newblock Google’s multilingual neural machine translation system: Enabling
  zero-shot translation.
\newblock \emph{Transactions of the Association for Computational Linguistics},
  5:339--351.

\bibitem[{Kasai et~al.(2019)Kasai, Qian, Gurajada, Li, and Popa}]{kasai2019low}
Jungo Kasai, Kun Qian, Sairam Gurajada, Yunyao Li, and Lucian Popa. 2019.
\newblock Low-resource deep entity resolution with transfer and active
  learning.
\newblock \emph{arXiv preprint arXiv:1906.08042}.

\bibitem[{Kim(2014)}]{kim2014convolutional}
Yoon Kim. 2014.
\newblock Convolutional neural networks for sentence classification.
\newblock \emph{arXiv preprint arXiv:1408.5882}.

\bibitem[{Lample et~al.(2016)Lample, Ballesteros, Subramanian, Kawakami, and
  Dyer}]{lample2016neural}
Guillaume Lample, Miguel Ballesteros, Sandeep Subramanian, Kazuya Kawakami, and
  Chris Dyer. 2016.
\newblock Neural architectures for named entity recognition.
\newblock In \emph{Proceedings of NAACL-HLT}, pages 260--270.

\bibitem[{Lample et~al.(2017)Lample, Conneau, Denoyer, and
  Ranzato}]{lample2017unsupervised}
Guillaume Lample, Alexis Conneau, Ludovic Denoyer, and Marc'Aurelio Ranzato.
  2017.
\newblock Unsupervised machine translation using monolingual corpora only.
\newblock \emph{arXiv preprint arXiv:1711.00043}.

\bibitem[{Liu et~al.(2018)Liu, Xu, Sinha, and Akkiraju}]{liu2018voice}
Xiaotong Liu, Anbang Xu, Vibha Sinha, and Rama Akkiraju. 2018.
\newblock Voice of customer: a tone-based analysis system for online user
  engagement.
\newblock In \emph{Extended Abstracts of the 2018 CHI Conference on Human
  Factors in Computing Systems}, page LBW001. ACM.

\bibitem[{Mikolov et~al.(2013)Mikolov, Le, and
  Sutskever}]{mikolov2013exploiting}
Tomas Mikolov, Quoc~V Le, and Ilya Sutskever. 2013.
\newblock Exploiting similarities among languages for machine translation.
\newblock \emph{arXiv preprint arXiv:1309.4168}.

\bibitem[{Peters et~al.(2018)Peters, Neumann, Iyyer, Gardner, Clark, Lee, and
  Zettlemoyer}]{peters2018deep}
Matthew~E Peters, Mark Neumann, Mohit Iyyer, Matt Gardner, Christopher Clark,
  Kenton Lee, and Luke Zettlemoyer. 2018.
\newblock Deep contextualized word representations.
\newblock \emph{arXiv preprint arXiv:1802.05365}.

\bibitem[{Radford et~al.(2018)Radford, Narasimhan, Salimans, and
  Sutskever}]{radford2018improving}
Alec Radford, Karthik Narasimhan, Tim Salimans, and Ilya Sutskever. 2018.
\newblock Improving language understanding by generative pre-training.
\newblock \emph{URL https://s3-us-west-2. amazonaws.
  com/openai-assets/researchcovers/languageunsupervised/language understanding
  paper. pdf}.

\bibitem[{Ramshaw and Marcus(1999)}]{ramshaw1999text}
Lance~A Ramshaw and Mitchell~P Marcus. 1999.
\newblock Text chunking using transformation-based learning.
\newblock In \emph{Natural language processing using very large corpora}, pages
  157--176. Springer.

\bibitem[{Tjong Kim~Sang(2002)}]{TjongKimSang:2002}
Erik~F. Tjong Kim~Sang. 2002.
\newblock \href {https://doi.org/10.3115/1118853.1118877} {Introduction to the
  conll-2002 shared task: Language-independent named entity recognition}.
\newblock In \emph{Proceedings of the 6th Conference on Natural Language
  Learning - Volume 20}, COLING-02, pages 1--4, Stroudsburg, PA, USA.
  Association for Computational Linguistics.

\bibitem[{Tjong Kim~Sang and De~Meulder(2003)}]{TjongKimSang:2003}
Erik~F. Tjong Kim~Sang and Fien De~Meulder. 2003.
\newblock \href {https://doi.org/10.3115/1119176.1119195} {Introduction to the
  conll-2003 shared task: Language-independent named entity recognition}.
\newblock In \emph{Proceedings of the Seventh Conference on Natural Language
  Learning at HLT-NAACL 2003 - Volume 4}, CONLL '03, pages 142--147,
  Stroudsburg, PA, USA. Association for Computational Linguistics.

\bibitem[{Vaswani et~al.(2017)Vaswani, Shazeer, Parmar, Uszkoreit, Jones,
  Gomez, Kaiser, and Polosukhin}]{vaswani2017attention}
Ashish Vaswani, Noam Shazeer, Niki Parmar, Jakob Uszkoreit, Llion Jones,
  Aidan~N Gomez, {\L}ukasz Kaiser, and Illia Polosukhin. 2017.
\newblock Attention is all you need.
\newblock In \emph{Advances in neural information processing systems}, pages
  5998--6008.

\bibitem[{Wu et~al.(2016)Wu, Schuster, Chen, Le, Norouzi, Macherey, Krikun,
  Cao, Gao, Macherey et~al.}]{wu2016google}
Yonghui Wu, Mike Schuster, Zhifeng Chen, Quoc~V Le, Mohammad Norouzi, Wolfgang
  Macherey, Maxim Krikun, Yuan Cao, Qin Gao, Klaus Macherey, et~al. 2016.
\newblock Google's neural machine translation system: Bridging the gap between
  human and machine translation.
\newblock \emph{arXiv preprint arXiv:1609.08144}.

\end{thebibliography}
